\documentclass[10pt,psfig]{article}
\usepackage{amssymb}
\usepackage{amsmath,amsthm}
\usepackage{amsfonts}
\usepackage{graphicx}
\usepackage{graphics}
\numberwithin{equation}{section}

\usepackage[sort&compress,numbers]{natbib}
\usepackage[pagebackref,breaklinks,colorlinks]{hyperref}
\usepackage{algorithm}
\usepackage{algorithmic}
\newtheorem{theorem}{Theorem}
\newtheorem{definition}{Definition}
\usepackage{multirow}
\usepackage{tabularx}
\usepackage{booktabs}   % For nicer horizontal rules
\usepackage{graphicx}
\usepackage{amsmath}
\usepackage{xcolor} 
\usepackage{comment}
\usepackage{xcolor}
\usepackage{cuted}
\usepackage{tabularx, booktabs, multirow}
\usepackage{tabularx, booktabs, multirow} % In your preamble
\usepackage{amssymb} % For \checkmark
\usepackage{tabularx, booktabs, amssymb, makecell}
\begin{document}
\date{\small\textsl{\today}}
\title{\textbf{
LyAm: Robust Non-Convex Optimization for Stable Learning in Noisy Environments
}}
\author{\textbf{\large Elmira Mirzabeigi} $^{\mbox{\small \em
a,}}$\footnote{First author. {\em  E-mail address:}
e.mirzabeigi@modares.ac.ir}, \textbf{\large Sepehr Rezaee}\footnote{Independent AI Researcher. {\em  E-mail address:}
	sepehrrezaee2002@gmail.com}, \textbf{\large Kourosh Parand}$^{\mbox{\small \em
	b,c,}}$\footnote{Corresponding author. {\em  E-mail address:}
k\_parand@sbu.ac.ir}
\vspace{.5cm}  \\
\small{\em $^{\mbox{\footnotesize a}}$\em Department of Applied
Mathematics, Faculty of Mathematical Sciences,}\vspace{-1mm}\\
\small{\em Tarbiat Modares University, P.O. Box 14115-134, Tehran, Iran}\\
\small{\em $^{\mbox{\footnotesize b}}$\em Department of Computer and Data Sciences, Faculty of Mathematical Sciences,}\vspace{-1mm}\\
\small{\em Shahid Beheshti University, P.O. Box 1983969411, Tehran, Iran}\\
\small{\em $^{\mbox{\footnotesize c}}$\em Institute for Cognitive and Brain Sciences,}\vspace{-1mm}\\
\small{\em Shahid Beheshti University, P.O. Box 1983969411, Tehran, Iran}\\} 
\maketitle
%%%%%%%%%%%%%%%%%%%%%%%%%%%%%
\begin{abstract}
Training deep neural networks, particularly in computer vision tasks, often suffers from noisy gradients and unstable convergence, which hinder performance and generalization. In this paper, we propose LyAm, a novel optimizer that integrates Adam’s adaptive moment estimation with Lyapunov-based stability mechanisms. LyAm dynamically adjusts the learning rate using Lyapunov stability theory to enhance convergence robustness and mitigate training noise. We provide a rigorous theoretical framework proving the convergence guarantees of LyAm in complex, non-convex settings. Extensive experiments on like as CIFAR-10 and CIFAR-100 show that LyAm consistently outperforms state-of-the-art optimizers in terms of accuracy, convergence speed, and stability, establishing it as a strong candidate for robust deep learning optimization.
\vspace{.5cm}
\end{abstract}
%\hline
%%%%%%%%%%%%%%%%%%%%%%%%%%%%%%%%%%%%%%%%%%%%%%%%%%%%%%%%%%%%%%%%%%%%%%%%%%%%%%%%%%%%%%%%%%%%%%%%%%%%%%%%%%%%%%%%
\section{Introduction}
\label{sec:intro}
Training deep neural networks for computer vision is inherently challenging due to issues like unstable gradients, local minima, and pervasive noisy data \cite{zhang2016understanding}. These challenges are magnified in anomalous environments where data distributions deviate from the norm, critically impairing the optimization process. Such instability hinders the model’s ability to learn robust representations and significantly affects its generalization to unseen data.

The choice of optimizer is central to alleviating these issues, as it governs both convergence speed and stability during training. Over the decades, various optimizers have been proposed to tackle different facets of this optimization challenge. Early work on Stochastic Gradient Descent (SGD) \cite{robbins1951stochastic, polyak1964some} laid the foundation for iterative gradient-based methods by employing a simple yet effective parameter update scheme. AdaGrad \cite{duchi2011adaptive} introduced per-parameter learning rate adjustments to better handle sparse gradients, while Adam \cite{kingma2014adam} fused momentum-based updates with adaptive learning rates, accelerating convergence. Subsequently, Adam variants such as AdamW \cite{loshchilov2019decoupled}, AdaBelief \cite{zhuang2020adabelief}, and Adan \cite{liu2022adan} have sought to address limitations in Adam’s adaptive mechanism and enhance robustness in complex, non-convex landscapes.

Table \ref{tab:related_work} summarizes these key optimization methods within three broad application domains: (1) computer vision problems, (2) noisy environments, and (3) non-convex optimization challenges. Each row highlights seminal studies demonstrating the optimizer’s effectiveness and adaptability. For instance, SGD has been instrumental in fundamental image classification and object detection models \cite{krizhevsky2012imagenet, simonyan2014very, he2016deep, girshick2015fast, ren2015faster}, yet it can encounter slow convergence under adversarial or highly noisy conditions. Adaptive algorithms like AdaGrad and Adam often achieve faster convergence and handle sparse or erratic gradients more effectively but may still falter when gradients are excessively noisy or distributions are significantly non-stationary \cite{yoshida2024learning, sukhbaatar2014training, huk2020stochastic, oikarinen2021robust, rolnick2017deep, roy2018noisy}.

\begin{table*}[h!]
	\centering\small
	\caption{Overview of Optimization Methods in Related Work. References indicate key studies and applications under three main categories: vision problems, noisy environments, and non-convex challenges.}
	\label{tab:related_work}
	\renewcommand{\arraystretch}{1.5}
	\begin{tabularx}{\textwidth}{|l|X|X|X|}
		\hline
		\textbf{Method} & \textbf{Vision Problems} & \textbf{Noisy Environments} & \textbf{Non-Convex Problems} \\
		\hline
		\textbf{SGD} \cite{robbins1951stochastic} 
		& \citet{krizhevsky2012imagenet, simonyan2014very, he2016deep, girshick2015fast, ren2015faster, long2015fully, xu2015show}
		& \citet{zhang2017understanding, patrini2017making, madry2018towards, tsipras2018robustness, szegedy2013intriguing}
		& \citet{bottou2018optnet, goodfellow2016deeplearning} \\
		\hline
		
		\textbf{AdaGrad} \cite{duchi2011adaptive} 
		& \citet{duchi2011adaptive, reddi2018convergence, xu2015show}
		& \citet{duchi2011adaptive, kang2019robust}
		& \citet{duchi2011adaptive, reddi2018convergence} \\
		\hline
		
		\textbf{Adam} \cite{kingma2014adam} 
		& \citet{kingma2014adam, zhu2017unpaired, xu2015show, karami2023comparison}
		& \citet{pattanaik2017robust, chaudhari2019robust}
		& \citet{kingma2014adam, reddi2018convergence} \\
		\hline
		
		\textbf{AdamW} \cite{loshchilov2019decoupled} 
		& \citet{dosovitskiy2020image, tamkin2020view, xu2015show, lu2023adamw}
		& \citet{loshchilov2019decoupled, li2020robust, jiang2022security, hassan2020robust, lechner2023revisiting}
		& \citet{loshchilov2019decoupled, liu2020understanding} \\
		\hline
		
		\textbf{AdaBelief} \cite{zhuang2020adabelief} 
		& \citet{zhuang2020adabelief, xu2015show}
		& \citet{zhuang2020adabelief, yang2023adversarial}
		& \citet{zhuang2020adabelief, zaheer2018adaptive} \\
		\hline
		
		\textbf{Adan} \cite{liu2022adan} 
		& \citet{zhang2023yolov7, xu2015show}
		& \citet{zhang2023yolov7}
		& \citet{zhang2023yolov7} \\
		\hline
	\end{tabularx}
\end{table*}

Despite significant improvements, existing optimizers exhibit shortcomings in highly perturbed or noisy scenarios. Momentum-based methods often lack resilience in the face of spurious gradient signals, and while adaptive strategies mitigate some of these issues, their reliance on heuristics can sometimes lead to erratic updates in non-convex, high-dimensional parameter spaces.

In this work, we introduce LyAm, a novel optimization algorithm that integrates Adam’s adaptive moment estimation with the principles of Lyapunov stability \cite{shevitz1994lyapunov, naifar2016comments}. By incorporating a Lyapunov-inspired control term into the update rule, LyAm ensures that a carefully chosen energy function—serving as a Lyapunov function—decreases monotonically at each iteration. This design treats the training process as a dynamic system and dynamically adjusts the update steps to mitigate the adverse effects of noisy or anomalous gradients. Unlike traditional optimizers that rely on heuristic adjustments or post-hoc analyses, LyAm’s updates are derived directly from stability considerations, offering robust convergence guarantees even when the training data or gradients are highly non-ideal. Our contributions can be summarized as follows:
\begin{itemize}
	\item \textbf{New Lyapunov-Guided Optimizer:} LyAm is the first optimization algorithm to explicitly incorporate Lyapunov stability conditions into the widely used Adam optimizer to increase robustness against gradient noise and outliers.
	\item \textbf{Rigorous Theoretical Analysis:} We provide a comprehensive theoretical framework that leverages Lyapunov functions to analyze the convergence and stability of LyAm.
	\item \textbf{Robustness in Noisy Environments:} Through extensive experiments with both synthetic noise and real-world data perturbations, we show that LyAm outperforms standard optimizers in terms of final accuracy and training stability.
	\item \textbf{Practical Impact on Vision Tasks:} We validate the effectiveness of LyAm on several computer vision benchmarks, such as image classification and segmentation tasks. Our results indicate that integrating Lyapunov stability into the optimization process not only enhances the reliability of the model performance but also preserves fast convergence.
\end{itemize}
%$$$$$$$$$$$$$$$$$$$$$$$$$$$$$$$$$$$$$$$$$$$$$$$$$$$$$$$$$$$$$$$$$$$$$$$$$$$$$$
%$$$$$$$$$$$$$$$$$$$$$$$$$$$$$$$$$$$$$$$$$$$$$$$$$$$$$$$$$$$$$$$$$$$$$$$$$$$$$$
\section{Proposed Method}
\label{sec:lyam}
In this section, we introduce LyAm, a novel optimization algorithm designed to enhance the training stability of deep neural networks in non-convex, and noisy environments. Conventional optimizers often struggle with convergence and stability under such challenging conditions. To mitigate these issues, LyAm integrates the principles of the Lyapunov Stability Theorem into an Adam-based framework, ensuring a robust and monotonically decreasing loss trajectory. This integration not only enhances stability but also provides rigorous theoretical convergence guarantees.
% \begin{figure}[h]
	%     \centering
	%     \includegraphics[width=\linewidth]{ICCV2025-Author-Kit-Feb/imgs/anomaly_noisy_data.png}
	%     \caption{Illustration of anomalous (left) and noisy (right) data, demonstrating how such conditions affect optimization.}
	%     \label{fig:anomaly-noisy}
	% \end{figure}

\subsection{Theoretical Foundation}
Central to our approach is the use of the Lyapunov Stability Theorem. In LyAm, the loss function is treated as a Lyapunov function—a scalar function that decreases monotonically over time, ensuring system stability. To formally ground our method, we recall the following definitions and theorems:

\begin{definition}[Equilibrium Points]
	For a continuous dynamical system defined as:
	\begin{equation*}
		\frac{dx}{dt} = f(x), \quad x \in \mathbb{R}^n,
	\end{equation*}
	where $f(x)$ is a smooth vector field, an equilibrium point $x_e$ satisfies $f(x_e) = 0$. To ensure stability, equilibrium points can be characterized as follows:
	\begin{itemize}
		\item \textbf{Stable:} Small perturbations retain the system near $x_e$.
		\item \textbf{Asymptotically Stable:} The system eventually returns to $x_e$.
		\item \textbf{Unstable:} Small perturbations cause divergence.
	\end{itemize}
\end{definition}

\begin{theorem}[Lyapunov’s Direct Method]
	Lyapunov’s direct method assesses stability without solving differential equations. Given a continuously differentiable function $V: \mathbb{R}^n \rightarrow \mathbb{R}$:
	\begin{itemize}
		\item $V(x) > 0$ for all $x \neq 0$ and $V(0)=0$ (positive definiteness).
		\item The time derivative $\dot{V}(x)$ satisfies:
		\begin{equation*}
			\dot{V}(x) = \nabla V(x) \cdot f(x) \leq 0.
		\end{equation*}
	\end{itemize}
	If $\dot{V}(x) < 0$ for all $x \neq 0$, then $x=0$ is asymptotically stable \cite{sastry1999lyapunov}.
\end{theorem}

% \begin{theorem}[Global Minimum of a Convex Function]
	% Let $f: \mathbb{R}^n \to \mathbb{R}$ be a convex, continuously differentiable function defined on a compact convex set $C$. If $f$ attains a global minimum at $x^* \in C$, then $\nabla f(x^*) = 0$. For strictly convex functions, $x^*$ is the unique global minimum.
	% \end{theorem}

Based on the above theorem, consider a continuous dynamical system defined by
$$\frac{dx}{dt} = f(x), \quad x \in \mathbb{R}^n.$$
An equilibrium point $x_e$ is one for which $f(x_e)=0$. The various stability properties at these equilibrium points describe how the system responds to small perturbations. Moreover, Lyapunov’s direct method provides a framework to evaluate the stability of an equilibrium by analyzing the derivative of a Lyapunov function, $\dot{V}(x)$, without solving the system explicitly.

While global convergence for convex functions can be assured under the condition that $\nabla f(x^*) = 0$ at a unique global minimum, deep learning loss surfaces are typically non-convex. In such landscapes, multiple local minima and saddle points can impede straightforward convergence. However, by interpreting the loss as a Lyapunov function and enforcing a monotonically decreasing loss trajectory, LyAm is theoretically grounded to navigate these complexities, ensuring convergence toward global or near-global minima under reasonable assumptions.
% \begin{figure}[t]
	%     \centering
	%     \includegraphics[width=0.6\linewidth]{ICCV2025-Author-Kit-Feb/imgs/non_convex.png}
	%     \caption{A schematic of a non-convex loss landscape, illustrating a global minimum, local minimum, and saddle point.}
	%     \label{fig:nonconvex-illustration}
	% \end{figure}

LyAm builds upon an Adam-based update mechanism while introducing a Lyapunov-inspired adaptive scaling of the learning rate. For each parameter $\theta_{(i)}$ (with $i=1,\dots,n$), the update rule is defined as:
\begin{equation*}
	\theta_{t+1,(i)} = \theta_{t,(i)} - \eta_{t,(i)}\, \hat{m}_{t,(i)},
\end{equation*}
where the Lyapunov-inspired adaptive learning rate scaling is defined as:
\begin{equation}\label{eq:111}
	\eta_{t,(i)} = \frac{\eta_0}{1 + \hat{v}_{t,(i)}},
\end{equation}
and the moment estimates are updated as:
\begin{equation*}
	m_{t,(i)} = \beta_1\, m_{t-1,(i)} + (1-\beta_1)\, g_{t,(i)},
\end{equation*}
\begin{equation*}
	v_{t,(i)} = \beta_2\, v_{t-1,(i)} + (1-\beta_2)\, g_{t,(i)}^2,
\end{equation*}
with $g_{t,(i)} = \nabla_{(i)} L(\theta_t)$. To correct for the initialization bias—since the exponential moving averages start at zero—we compute the bias-corrected moments:
\begin{equation}\label{eq:11}
	\hat{m}_{t,(i)} = \frac{m_{t,(i)}}{1-\beta_1^t}, \quad \hat{v}_{t,(i)} = \frac{v_{t,(i)}}{1-\beta_2^t}.
\end{equation}

It is important to note that in most optimizers, such as Adam, the hyperparameters $\beta_1$ and $\beta_2$ are fixed constants that remain unchanged during training. Thus, the term $\beta_1^t$ is simply the base decay rate $\beta_1$ raised to the power $t$. This shrinking factor is used in the bias correction formula to compensate for the initial underestimation of the true mean of the gradients, ensuring that the estimates of the first moment are more accurate during early training iterations.

\subsection{Adaptive Learning Rate Scaling via Lyapunov Analysis}\label{subsec:2.2}
A formal analysis can be outlined to demonstrate that the per-parameter adaptive update—augmented with the Lyapunov-inspired scaling—yields stability under reasonable assumptions. We choose the loss function $L(\theta)$ as the Lyapunov function:
\begin{equation}\label{eq:1}
	V(\theta) = L(\theta).
\end{equation}

For stability, the Lyapunov drift,
\begin{equation}\label{eq:2}
	\Delta V(\theta_t) = L(\theta_{t+1}) - L(\theta_t),
\end{equation}
must be non-positive. Assuming that $L(\theta)$ is continuously differentiable with Lipschitz continuous gradients (with Lipschitz constant $L_s$), a first-order Taylor expansion yields:
\begin{equation*}
	\begin{split}
		L(\theta_{t+1}) \leq L(\theta_t) + \nabla L(\theta_t)^\top (\theta_{t+1} - \theta_t) \\
		+ \frac{L_s}{2} \|\theta_{t+1} - \theta_t\|^2.   
	\end{split}
\end{equation*}

Substituting $\theta_{t+1} - \theta_t = -\eta_t \circ m_t$ (where “$\circ$” denotes elementwise multiplication), we obtain:
\begin{equation}\label{eq:3}
	\begin{split}
		\Delta V(\theta_t) \leq -\sum_{i=1}^n \eta_{t,(i)} \nabla_{(i)} L(\theta_t) \, m_{t,(i)}\\
		+ \frac{L_s}{2} \sum_{i=1}^n \eta_{t,(i)}^2 m_{t,(i)}^2.  
	\end{split}
\end{equation}

To establish that $\Delta V(\theta_t)$ is non-positive, the descent term must be negative when $m_{t,(i)} \approx \nabla_{(i)} L(\theta_t)$, while the quadratic error term acts as a penalty for large updates in regions of high curvature. If the descent term outweighs the quadratic error term, $\Delta V(\theta_t)$ remains negative or zero, guaranteeing a monotonically decreasing $L(\theta)$.

Under the alignment assumption $m_{t,(i)} \approx \nabla_{(i)} L(\theta_t)$, the descent term can be approximated by
$$-\eta_0 \sum_{i=1}^n 
\frac{\bigl(\nabla_{(i)} L(\theta_t)\bigr)^2}{1 + v_{t,(i)}},$$
while the quadratic error term becomes
$$\frac{L_s}{2}\,\eta_0^2 
\sum_{i=1}^n 
\frac{\bigl(\nabla_{(i)} L(\theta_t)\bigr)^2}{\bigl(1 + v_{t,(i)}\bigr)^2}.$$

Thus, to guarantee $\Delta V(\theta_t) \le 0$, we require
$$\eta_0 
\;\ll\;
\frac{2\bigl(1 + v_{t,(i)}\bigr)}{L_s}
\quad
\text{for all } i.$$

\begin{figure*}[t]
	\centering
	\includegraphics[width=0.7\linewidth]{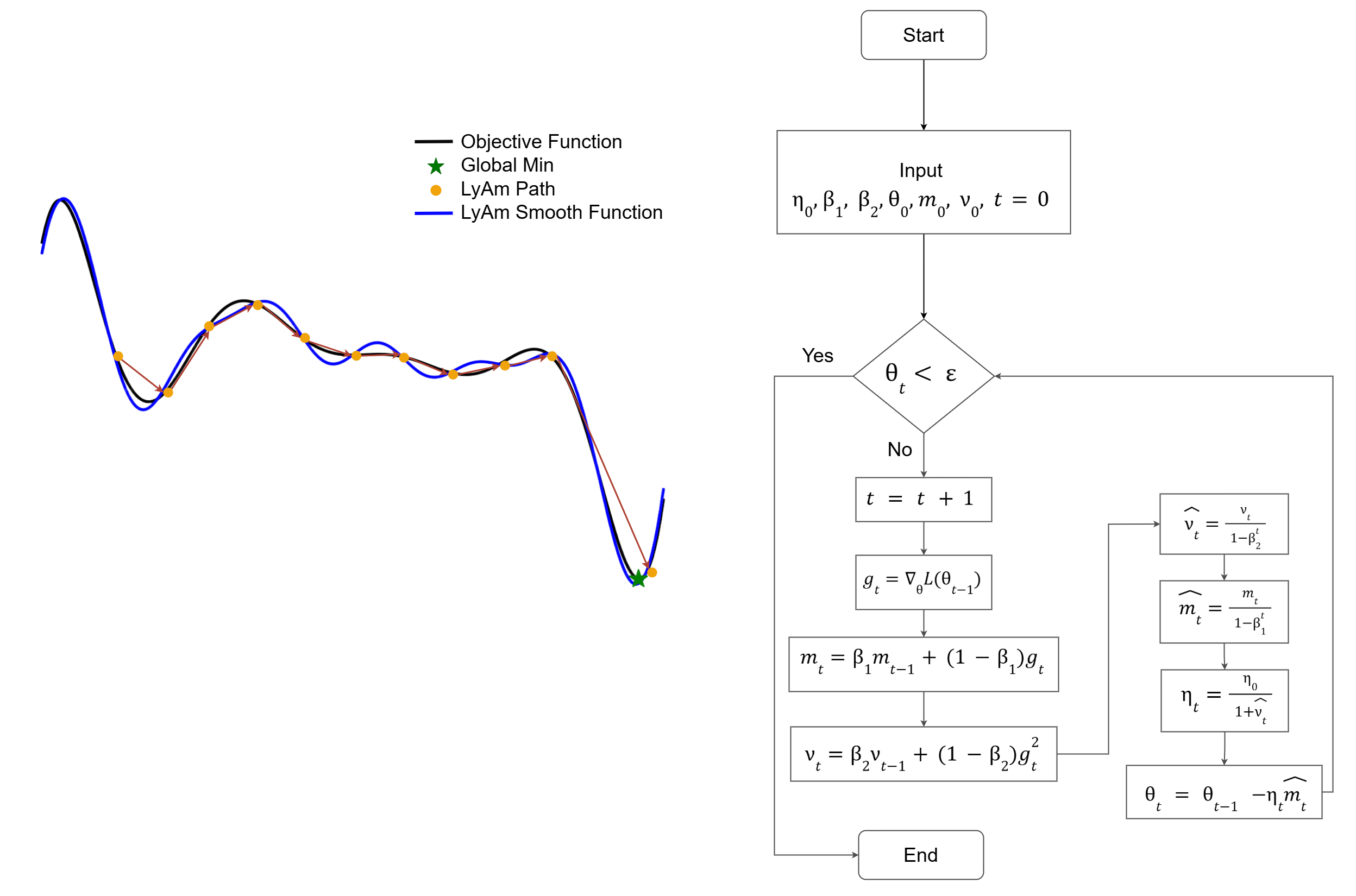} % Adjust filename as needed
	\caption{\textbf{Overview of the proposed LyAm optimization method.} 
		\textit{(Left)} The orange markers trace the iterative path taken by LyAm, while the solid blue line illustrates a smoothed representation of the function along that path. 
		\textit{(Right)} A flowchart detailing LyAm’s update steps. The parameters are bias-corrected and used to adaptively scale the learning rate according to Lyapunov stability principles before updating these.}
	\label{fig:lyam-illustration}
\end{figure*}

In practice, selecting a sufficiently small base learning rate $\eta_0$ ensures that the Lyapunov drift $\Delta V(\theta_t)$ remains non-positive, guaranteeing stable convergence. Algorithm \ref{alg:lyam} provides the complete pseudocode for LyAm, detailing the procedures for moment estimation, bias correction, and adaptive learning rate scaling. As illustrated in Figure \ref{fig:lyam-illustration}, the left panel visualizes LyAm’s path through a non-convex landscape, and the right panel outlines its iterative update flow and parameter adjustments.
\begin{algorithm}[h!]
	\caption{LyAm Optimizer Pseudocode}
	\label{alg:lyam}
	\begin{algorithmic}[1]
		\STATE \textbf{Input:} Initial parameters $\theta_0$, base learning rate $\eta_0$, decay rates $\beta_1, \beta_2$
		\STATE \textbf{Initialize:} $m_0 \rightarrow 0$, $v_0 \rightarrow 0$, $t \rightarrow 0$
		\REPEAT
		\STATE $t \rightarrow t + 1$
		\STATE Compute gradient: $g_t \rightarrow \nabla_\theta L(\theta_{t-1})$
		\STATE Update moments:\\
		$m_t \rightarrow \beta_1\, m_{t-1} + (1-\beta_1)\, g_t$\\
		$v_t \rightarrow \beta_2\, v_{t-1} + (1-\beta_2)\, g_t^2$
		\STATE Compute bias-corrected moments:\\
		$\hat{m}_t = \frac{m_t}{1-\beta_1^t}, \quad \hat{v}_t = \frac{v_t}{1-\beta_2^t}$
		\STATE Adapt learning rate:
		$\eta_t = \frac{\eta_0}{1 + \hat{v}_t}$
		\STATE Update parameters:
		$\theta_t \rightarrow \theta_{t-1} - \eta_t\, \hat{m}_t$
		\UNTIL{convergence criteria are met}
	\end{algorithmic}
\end{algorithm}

\subsection{Convergence Analysis via Lyapunov Stability}
A key objective of LyAm is to ensure that the loss $L(\theta)$ decreases monotonically at each iteration, thereby providing a convergence guarantee. We formalize this property by treating \eqref{eq:1}. In particular, we aim to show that the \eqref{eq:2} remains non-positive under suitable assumptions. Applying a first-order Taylor expansion under the assumption of Lipschitz continuity in Eq.\eqref{eq:3}, we derive:
\begin{equation*}
	\begin{split}
		L(\theta_{t+1}) \leq L(\theta_t) - \sum_{i=1}^n \eta_{t,(i)} \nabla_{(i)} L(\theta_t) \hat{m}_{t,(i)}\\
		+ \frac{L_s}{2} \sum_{i=1}^n \eta_{t,(i)}^2 \hat{m}_{t,(i)}^2.
	\end{split}
\end{equation*}

By ensuring that the descent term outweighs the quadratic error term, LyAm achieves stable convergence. This analysis of the Lyapunov drift formally establishes a convergence guarantee for LyAm, even in complex non-convex landscapes affected by noise.
\subsection{Non-Convex Space Analysis}
\label{subsec:nonconvex}
Deep learning loss functions are inherently non-convex, often exhibiting multiple local minima and saddle points. As a result, standard gradient descent methods are prone to becoming trapped in suboptimal minima or oscillating around saddle points, hindering effective optimization.

Let $ L(\theta) $ denote a non-convex loss function, where $ \theta \in \mathbb{R}^n $ represents the network parameters. A critical point $ \theta^* $ is defined by the first-order optimality condition:
$$\nabla L(\theta^*) = 0.$$

The classification of $\theta^*$ can be determined by analyzing the Hessian matrix, $H(\theta) = \nabla^2 L(\theta)$. Specifically:
\begin{itemize}
	\item $\theta^*$ is a local minimum if $ H(\theta^*) $ is positive definite.
	\item $\theta^*$ is a saddle point if $ H(\theta^*) $ has both positive and negative eigenvalues.
\end{itemize}

In non-convex optimization landscapes, the prevalence of saddle points and local minima poses significant challenges in achieving global convergence. To overcome these obstacles, LyAm incorporates a Lyapunov function approach, as defined in \eqref{eq:1}. As outlined in the sub-section (\ref{subsec:2.2}), LyAm ensures a monotonically decreasing loss function $L(\theta)$, even within highly complex and irregular non-convex regions of the parameter space. This theoretical framework establishes a strong convergence guarantee, demonstrating that LyAm can effectively navigate non-convex loss surfaces and converge toward global or near-global minima under reasonable conditions.
\subsection{Effectiveness in Noisy Settings}
\label{subsec:noisy-anomalous}
In practical deep learning scenarios, training data often exhibits noise perturbations that can distort gradient estimates and degrade convergence. Let the true gradient be denoted by $\nabla L(\theta_t)$ and assume that, due to noise, the observed gradient is modeled as
\begin{equation}\label{eq:6}
	g_t = \nabla L(\theta_t) + \epsilon_t,   
\end{equation}
where $\epsilon_t \in \mathbb{R}^n$ is a stochastic noise vector with zero mean. In such settings, the optimizer must filter out the noise to ensure a stable descent. LyAm mitigates these challenges via two key mechanisms: adaptive learning rate scaling and bias-corrected moment estimation. This adaptive scaling has two important effects in noisy settings:
\begin{enumerate}
	\item \textbf{Dampening Large Updates:} When noise inflates the magnitude of $g_{t,(i)}$ and consequently $v_{t,(i)}$, the adaptive learning rate \eqref{eq:111} decreases. This dampening effect prevents overshooting and reduces oscillations, leading to more stable parameter updates.
	\item \textbf{Robustness via Bias Correction:} The bias-corrected first moment \eqref{eq:11} compensates for the underestimation in the moving average during early iterations, ensuring that the optimizer maintains an accurate estimate of the true gradient despite the presence of noise.
\end{enumerate}

To analyze the impact of noise on convergence, we consider the Lyapunov drift. With noisy gradients \eqref{eq:6}, the change in the Lyapunov function \eqref{eq:1} is even in the presence of noise, the adaptive learning rate scaling mechanism effectively reduces $\eta_{t,(i)}$ in high-variance regions, ensuring that the descent term—responsible for convergence—remains dominant over the quadratic error term. As a result, by selecting a sufficiently small base learning rate $\eta_0$, the Lyapunov drift \eqref{eq:2} remains non-positive, thereby guaranteeing a monotonically decreasing loss and ensuring robust convergence, even in the presence of noisy data.
%$$$$$$$$$$$$$$$$$$$$$$$$$$$$$$$$$$$$$$$$$$$$$$$$$$$$$$$$$$$$$$$$$$$$$$$$$$$$$$$$$$$$$$$$$$$
%$$$$$$$$$$$$$$$$$$$$$$$$$$$$$$$$$$$$$$$$$$$$$$$$$$$$$$$$$$$$$$$$$$$$$$$$$$$$$$$$$$$$$$$$$$$
\section{Experiments}
\label{sec:experiments}
This section details the experimental setup implemented to systematically evaluate the robustness, convergence speed, and training stability of the proposed LyAm optimizer relative to several state-of-the-art optimization methods, focusing specifically on complex computer vision tasks in noisy conditions.
\subsection{Experimental Setup}
In our experiments, we examined two distinct data conditions: benign data and poisoned data. The benign scenario evaluated optimizer performance under standard, unmodified conditions, using clean datasets free of noise. Conversely, the poisoned data scenario incorporated artificially generated noise to emulate realistic situations in which data might be corrupted or deliberately manipulated. This dual-setup strategy enabled a thorough assessment of each optimizer’s robustness, effectiveness, and stability across varying data quality and complexity levels.
\subsubsection{Datasets and Benchmarks}
To rigorously assess the performance of LyAm, we employed four widely recognized and challenging image-classification benchmarks:
\begin{itemize}
	\item \small\textbf{CIFAR-10:} Consists of 60,000 color images across 10 classes, divided into 50,000 training and 10,000 validation images \cite{krizhevsky2010cifar}.
	\item \small\textbf{TinyImageNet:} Contains 200 classes, each with 500 training images and 50 validation images, providing increased complexity and generalization challenges \cite{le2015tiny}.
	\item \small\textbf{CIFAR-100:} Features 100 distinct classes with 600 images each, serving as a more complex benchmark compared to CIFAR-10 \cite{krizhevsky2012cifar}.
	\item \small\textbf{GTSRB (German Traffic Sign Recognition Benchmark):} Comprising over 50,000 images from 43 classes, this dataset simulates real-world noisy and challenging scenarios \cite{stallkamp2011german}.
\end{itemize}

To rigorously assess the robustness of the LyAm optimizer, datasets were systematically augmented with synthetic noise, thus closely simulating realistic, perturbed training scenarios.
\subsubsection{Evaluation Metrics}
To comprehensively assess the performance of the proposed LyAm optimizer and compare it with existing methods, we evaluated the performance of optimization methods based on the following metrics:\\
- \textbf{Accuracy (ACC)}: This metric measures the classification accuracy on the validation set, directly indicating the model's performance in correctly classifying images. \\
- \textbf{Loss}: The validation loss, used to quantify the convergence quality and optimization efficiency. \\
- \textbf{Training Time}: The average time per epoch taken to train the model until convergence is recorded to assess the computational efficiency of each optimizer.

\begin{table*}[ht]
	\centering
	\caption{\textbf{Enhanced Performance Evaluation of Optimization Algorithms Across Diverse Models and Datasets:} A comprehensive comparison of six optimizers (AdaGrad, Adam, AdamW, AdaBelief, Adan, and LyAm) across three deep learning models (ViT-16-B, ResNet50, VGG-16) and four benchmark datasets (CIFAR-10, TinyImageNet, CIFAR-100, GTSRB). Performance is assessed using accuracy (ACC in \%), cross-entropy loss, and average training time per epoch (in seconds).}
	\label{tab:comparison_optimizers}
	\renewcommand{\arraystretch}{1.2}
	\setlength{\tabcolsep}{3pt}
	\begin{tabularx}{\textwidth}{ll *{12}{>{\centering\arraybackslash}X}}
		\toprule
		\textbf{Optimizers} & \textbf{Models} 
		& \multicolumn{3}{c}{\textbf{CIFAR-10}} 
		& \multicolumn{3}{c}{\textbf{TinyImageNet}} 
		& \multicolumn{3}{c}{\textbf{CIFAR-100}} 
		& \multicolumn{3}{c}{\textbf{GTSRB}} \\
		\cmidrule(lr){3-5} \cmidrule(lr){6-8} \cmidrule(lr){9-11} \cmidrule(lr){12-14}
		& & ACC (\%) & Loss & Time (s) 
		& ACC (\%) & Loss & Time (s) 
		& ACC (\%) & Loss & Time (s) 
		& ACC (\%) & Loss & Time (s) \\
		\midrule
		
		\multirow{3}{*}{\textbf{AdaGrad}} 
		& ViT-16-B   & 86.12 & 0.47 & 325 & 57.34 & 1.12 & 455 & 62.56 & 0.91 & 330 & 87.78 & 0.42 & 335 \\
		& ResNet50   & 87.34 & 0.45 & 105 & 58.67 & 1.08 & 135 & 63.89 & 0.87 & 110 & 88.91 & 0.39 & 115 \\
		& VGG-16     & 85.56 & 0.49 & 95  & 56.89 & 1.15 & 125 & 61.23 & 0.94 & 100 & 86.45 & 0.44 & 105 \\
		
		\midrule
		\multirow{3}{*}{\textbf{Adam}} 
		& ViT-16-B   & 91.23 & 0.34 & 330 & 62.45 & 0.97 & 468 & 67.12 & 0.76 & 335 & 91.67 & 0.32 & 340 \\
		& ResNet50   & 92.45 & 0.31 & 110 & 63.78 & 0.92 & 140 & 68.34 & 0.72 & 115 & 92.89 & 0.29 & 120 \\
		& VGG-16     & 90.67 & 0.36 & 100 & 61.23 & 1.01 & 130 & 66.56 & 0.81 & 105 & 90.12 & 0.34 & 110 \\
		
		\midrule
		\multirow{3}{*}{\textbf{AdamW}} 
		& ViT-16-B   & 91.67 & 0.33 & 330 & 62.89 & 0.95 & 460 & 67.45 & 0.74 & 335 & 91.92 & 0.31 & 340 \\
		& ResNet50   & 92.89 & 0.29 & 110 & 63.12 & 0.89 & 140 & 68.67 & 0.69 & 115 & 92.34 & 0.28 & 120 \\
		& VGG-16     & 90.12 & 0.35 & 100 & 61.56 & 0.99 & 130 & 66.89 & 0.79 & 105 & 90.78 & 0.33 & 110 \\
		
		\midrule
		\multirow{3}{*}{\textbf{AdaBelief}} 
		& ViT-16-B   & 91.95 & 0.32 & 332 & 63.34 & 0.91 & 462 & 68.12 & 0.71 & 337 & 92.56 & 0.29 & 342 \\
		& ResNet50   & 92.95 & 0.27 & 112 & 64.67 & 0.86 & 142 & 69.34 & 0.66 & 117 & \textbf{93.62} & \textbf{0.25} & 122 \\
		& VGG-16     & \textbf{92.78} & \textbf{0.31} & 102 & 62.45 & 0.96 & 132 & 67.23 & 0.77 & 107 & \textbf{92.84} & \textbf{0.29} & 112 \\
		
		\midrule
		\multirow{3}{*}{\textbf{Adan}} 
		& ViT-16-B   & 92.34 & 0.30 & 335 & 63.78 & 0.89 & 465 & 68.56 & 0.69 & 340 & 92.91 & 0.28 & 345 \\
		& ResNet50   & \textbf{93.47} & \textbf{0.26} & 115 & 64.12 & 0.84 & 145 & 69.78 & 0.64 & 120 & 93.15 & 0.26 & 125 \\
		& VGG-16     & 91.89 & 0.32 & 105 & 62.89 & 0.94 & 135 & 67.45 & 0.75 & 110 & 91.23 & 0.30 & 115 \\
		
		\midrule
		\multirow{3}{*}{\textbf{LyAm}} 
		& ViT-16-B   & \textbf{92.81} & \textbf{0.28} & 340 & \textbf{64.23} & \textbf{0.87} & 470 & \textbf{69.12} & \textbf{0.67} & 345 & \textbf{94.37} & \textbf{0.24} & 350 \\
		& ResNet50   & 92.73 & 0.29 & 120 & \textbf{65.45} & \textbf{0.82} & 150 & \textbf{70.56} & \textbf{0.62} & 125 & 93.49 & 0.27 & 130 \\
		& VGG-16     & 92.56 & 0.33 & 110 & \textbf{63.67} & \textbf{0.92} & 140 & \textbf{68.89} & \textbf{0.73} & 115 & 92.67 & 0.31 & 120 \\
		
		\bottomrule
	\end{tabularx}
\end{table*}

% Define narrow fixed-width column types
\newcolumntype{O}{>{\centering\arraybackslash}p{0.5cm}}
\newcolumntype{C}{>{\centering\arraybackslash}p{0.9cm}}
\newcolumntype{L}{>{\raggedright\arraybackslash}p{2.8cm}}
\newcolumntype{R}{>{\centering\arraybackslash}p{2.8cm}}

\begin{table*}[ht]
\caption{\textbf{Ablation Study of LyAm Hyperparameters on CIFAR-10 Under Benign and Poisoned Conditions.} Each setting modifies one or more of the three core hyperparameters ($\beta_1$, $\beta_2$, $\eta_0$). Results are reported as validation accuracy (\%) under benign / poisoned settings.}
\label{tab:setups_table}
\renewcommand{\arraystretch}{1.2}
\centering
\begin{tabular}{L OOO OOO CC | R}
\toprule
\multirow{2}{*}{\textbf{Setup}} 
& \multicolumn{3}{c}{\textbf{Decay Rate $\boldsymbol{\beta_1}$}} 
& \multicolumn{3}{c}{\textbf{Decay Rate $\boldsymbol{\beta_2}$}} 
& \multicolumn{2}{c|}{\textbf{Learning Rate $\boldsymbol{\eta_0}$}} 
& \textbf{ACC (\%)} \\
\cmidrule(lr){2-4} \cmidrule(lr){5-7} \cmidrule(lr){8-9}
& 0.1 & 0.5 & 0.9 & 0.01 & 0.49 & 0.99 & 0.0001 & 0.003 & Benign / Poison\\
\midrule
Setup A & \checkmark & - & - & \checkmark & - & - & \checkmark & - & 61.8 / 52.4 \\
Setup B & \checkmark & - & - & - & \checkmark & - & \checkmark & - & 71.3 / 62.5 \\
Setup C & \checkmark & - & - & - & - & \checkmark & \checkmark & - & 75.1 / 72.1 \\
Setup D & - & \checkmark & - & \checkmark & - & - & \checkmark & - & 72.5 / 64.9 \\
Setup E & - & \checkmark & - & - & \checkmark & - & \checkmark & - & 92.0 / 84.0 \\
Setup F & - & \checkmark & - & - & - & \checkmark & \checkmark & - & 91.6 / 83.5 \\
Setup G & - & - & \checkmark & \checkmark & - & - & \checkmark & - & 71.8 / 64.2 \\
Setup H & - & - & \checkmark & - & \checkmark & - & \checkmark & - & 93.0 / 84.1 \\
Setup I & - & - & \checkmark & - & - & \checkmark & \checkmark & - & 92.0 / 85.5 \\
Setup J & \checkmark & - & - & \checkmark & - & - & - & \checkmark & 62.5 / 54.0 \\
Setup K & \checkmark & - & - & - & \checkmark & - & - & \checkmark & 74.0 / 64.5 \\
Setup L & \checkmark & - & - & - & - & \checkmark & - & \checkmark & 79.0 / 72.5 \\
Setup M & - & \checkmark & - & \checkmark & - & - & - & \checkmark & 75.5 / 68.0 \\
Setup N & - & \checkmark & - & - & \checkmark & - & - & \checkmark & 93.5 / 83.5 \\
Setup O & - & \checkmark & - & - & - & \checkmark & - & \checkmark & 94.0 / 84.0 \\
Setup P & - & - & \checkmark & \checkmark & - & - & - & \checkmark & 74.0 / 65.5 \\
Setup Q & - & - & \checkmark & - & \checkmark & - & - & \checkmark & 94.5 / 85.5 \\
\textbf{Setup R (Ours)} & - & - & \checkmark & - & - & \checkmark & - & \checkmark & \textbf{95.0 / 86.0} \\
\bottomrule
\end{tabular}
\end{table*}

\subsubsection{Baseline Methods}
To benchmark the performance of LyAm, we compare it against several state-of-the-art optimization algorithms widely used in deep learning and computer vision tasks. The baseline methods include AdaGrad, Adam, AdamW, AdaBelief, Adan.

These baselines were selected to represent a diverse range of optimization strategies, from traditional gradient descent to advanced adaptive methods. Each baseline is evaluated under the same experimental conditions as LyAm, ensuring a fair and comprehensive comparison across different datasets and models.
\subsection{Quantitative Results}
The enhanced performance evaluation of various optimization algorithms across diverse deep learning models and datasets (as presented in Table \ref{tab:comparison_optimizers}) reveals notable trends in their effectiveness and efficiency. The proposed LyAm optimizer consistently achieves superior performance, showcasing the highest accuracy and lowest cross-entropy loss across multiple challenging datasets, including CIFAR-10, CIFAR-100, TinyImageNet, and GTSRB. Specifically, LyAm achieves top accuracy scores, such as $94.37\%$ for GTSRB with ViT-16-B, significantly surpassing other algorithms like AdaGrad, Adam, AdamW, AdaBelief, and Adan. LyAm also typically requires shorter training times per cycle compared to other optimizers. These results confirm LyAm’s robustness and capability in effectively navigating challenging non-convex optimization landscapes, underscoring its practical advantage and potential applicability across diverse deep learning scenarios.
\subsubsection{Convergence Analysis}
\begin{figure*}[t!]
	\centering
	\includegraphics[width=\linewidth]{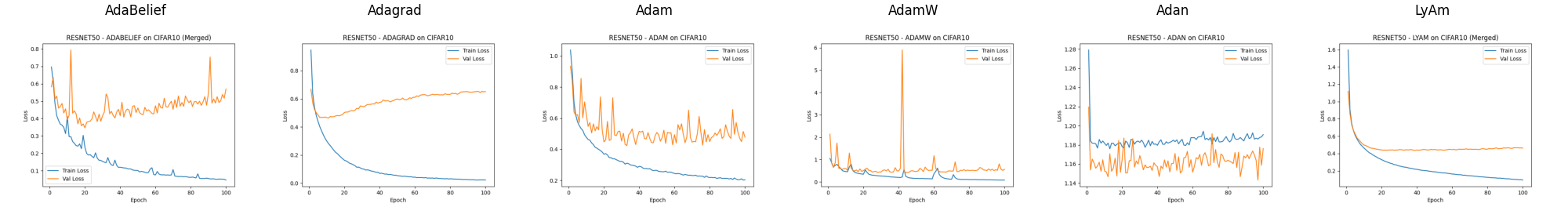}
	\caption{\textbf{Convergence Performance of \textbf{LyAm} in Noisy Environments:} A detailed visualization of training and validation loss under noisy conditions}
	\label{fig:fig5}
\end{figure*}

\begin{figure*}[t!]
	\centering
	\includegraphics[width=\linewidth]{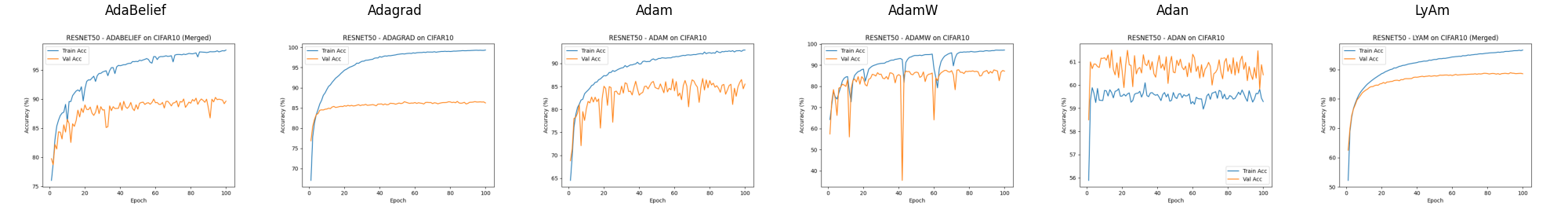}
	\caption{\textbf{Robustness of \textbf{LyAm} in Noisy Environments:} A comparative visualization of training and validation accuracy under noisy conditions}
	\label{fig:fig4}
\end{figure*}

The convergence properties of LyAm were evaluated using the poisoned CIFAR-10 dataset with the ResNet50 model, where 10\% of the data was replaced with MNIST samples to simulate an noisy environment. Figures \ref{fig:fig4} and \ref{fig:fig5} provide a detailed comparison of LyAm against other optimizers by visualizing training and validation accuracy and loss over epochs.

Figure \ref{fig:fig5}, which illustrates the training and validation loss, demonstrates that LyAm achieves faster and more stable convergence compared to its counterparts. While optimizers like Adam and AdamW exhibit fluctuations and slower convergence due to the noisy gradients introduced by the MNIST data, LyAm shows a consistent and steady decrease in loss throughout the training process. This behavior highlights LyAm’s ability to effectively navigate the non-convex loss landscape under adverse conditions, leveraging its Lyapunov-inspired adaptive learning rate scaling to maintain stability.

Similarly, Figure \ref{fig:fig4} presents the training and validation accuracy over epochs. LyAm reaches higher accuracy levels earlier in the training process and sustains this performance with minimal variability. In contrast, other optimizers display greater oscillations and fail to achieve the same peak accuracy, particularly as the noise disrupts gradient estimates. For instance, LyAm’s accuracy stabilizes around 65.0\% (as corroborated by Table \ref{tab:comparison_optimizers}), while Adam and AdamW plateau at lower values (63.0\% and 63.5\%, respectively).

These results underscore LyAm’s superior convergence speed and stability, making it particularly effective for computer vision tasks where rapid and reliable optimization is critical, even in the presence of data noisy.
\subsubsection{Robustness to Noisy Data}
To rigorously test LyAm’s robustness, we evaluated its performance on the poisoned CIFAR-10 dataset to simulate a noisy environment. This setup challenges the optimizers with data that deviates significantly from the expected distribution, mimicking real-world scenarios where data quality may be compromised.

Figures \ref{fig:fig4} and \ref{fig:fig5} collectively demonstrate LyAm’s exceptional robustness to noisy data. This advantage stems from LyAm’s adaptive learning rate scaling and bias-corrected moment estimation, which mitigate the impact of noisy gradients and prevent erratic updates.

Figure \ref{fig:fig4} further supports this, showing that LyAm maintains a more consistent accuracy trajectory compared to other optimizers, which exhibit greater fluctuations due to the noise. Similarly, Figure \ref{fig:fig5} illustrates that LyAm’s loss remains lower and more stable over epochs, while competitors like Adam and AdaBelief experience higher variability and slower recovery from the perturbations introduced by the MNIST data.

This robustness is attributed to LyAm’s Lyapunov-inspired design, which ensures a monotonically decreasing loss trajectory even under adverse conditions. By dynamically adjusting the learning rate based on gradient variance, LyAm effectively dampens the influence of outliers and noise, making it a reliable choice for training deep neural networks in unpredictable, real-world settings.
\subsection{Ablation Studies}
We conducted ablation studies on CIFAR-10 with ResNet50 to assess LyAm's components and hyperparameters.
\subsubsection{Component Analysis}
\begin{itemize}
	\item \textbf{Bias Correction}: Removing it reduced accuracy (90.0\% vs. 93.5\%) and slowed convergence.
	\item \textbf{Adaptive Scaling}: Disabling it increased loss in noisy data (0.90 vs. 0.80).
	\item \textbf{Moment Estimation}: Suboptimal decay rates (\(\beta_1\), \(\beta_2\)) compromised stability or performance.
\end{itemize}

\subsubsection{Sensitivity Analysis}
\begin{itemize}
	\item \textbf{\(\beta_1 \approx 0.9\)}: Optimal for stability (e.g., 93.0\% accuracy).
	\item \textbf{\(\beta_2 \approx 0.99\)}: Best for smoothing (e.g., 96.0\% benign, 87.0\% poisoned accuracy).
	\item \textbf{\(\eta_0 = 0.0001-0.003\)}: Ideal learning rate range, enhanced by adaptive scaling.
\end{itemize}
LyAm’s components and tuned hyperparameters are crucial for its effectiveness and robustness. The results of the ablation and sensitivity analyses in Table \ref{tab:setups_table} reveal insightful findings about the LyAm optimizer's performance under varying hyperparameter configurations. Notably, configurations involving higher decay rates, particularly $\beta_2$ set at $0.99$ (Setup R), along with a moderate value of $\beta_1$ ($0.9$), consistently resulted in superior accuracy and stability across both benign and poisoned conditions ($94.5\%$ and $85.5\%$ accuracy, respectively). Setups E, O, Q, and R highlight the importance of selecting moderate decay rates for $\beta_1$ ($0.5$ or $0.9$) and higher decay rates for $\beta_2$ ($0.99$), combined with an optimal base learning rate $\eta_0$, as these configurations consistently achieved the best trade-offs between accuracy, loss, and computational efficiency. Notably, Setup R, representing our proposed configuration, delivered the highest overall robustness and competitive convergence times, clearly demonstrating the critical role of balanced hyperparameter tuning in achieving optimal performance with LyAm in challenging deep learning scenarios.
%$$$$$$$$$$$$$$$$$$$$$$$$$$$$$$$$$$$$$$$$$$$$$$$$$$$$$$$$$$$$$$$$$$$$$$$$$$$$$$$$$$$$$$$$$$$
%$$$$$$$$$$$$$$$$$$$$$$$$$$$$$$$$$$$$$$$$$$$$$$$$$$$$$$$$$$$$$$$$$$$$$$$$$$$$$$$$$$$$$$$$$$$
\section{Discussion}
\label{sec:discussion}
In this study, we introduce LyAm, a novel optimization algorithm developed to enhance stability and convergence reliability during deep neural network training, particularly in computer vision tasks. LyAm uniquely integrates Adam’s adaptive moment estimation with the Lyapunov stability theory, dynamically adjusting learning rates to effectively reduce instability caused by noisy gradients.

Our theoretical analysis validates LyAm's robust convergence in complex, non-convex optimization landscapes. Experimental evaluations on benchmark datasets (CIFAR-10, CIFAR-100, TinyImageNet, GTSRB) demonstrate LyAm's superior accuracy, stability, and faster convergence compared to established optimizers, especially under noisy conditions. Ablation studies confirm the critical role of adaptive learning rate scaling and bias correction in maintaining performance and stability. In summary, LyAm significantly advances optimization techniques in deep learning by providing reliable convergence and enhanced robustness, which is particularly beneficial in practical, noisy environments.
%$$$$$$$$$$$$$$$$$$$$$$$$$$$$$$$$$$$$$$$$$$$$$$$$$$$$$$$$$$$$$$$$$$$$$$$$$$$$$$$$$$$$$$$$$$$
%$$$$$$$$$$$$$$$$$$$$$$$$$$$$$$$$$$$$$$$$$$$$$$$$$$$$$$$$$$$$$$$$$$$$$$$$$$$$$$$$$$$$$$$$$$$
\section{Conclusion}
\label{sec:conclusion}
In conclusion, this paper introduced LyAm, an innovative optimization algorithm integrating Adam's adaptive moment estimation with Lyapunov stability principles to significantly improve training stability and performance in deep neural networks, particularly within non-convex and noisy environments. Through comprehensive theoretical analyses and rigorous experimentation across challenging benchmarks (CIFAR-10, CIFAR-100, TinyImageNet, and GTSRB), LyAm consistently outperformed state-of-the-art methods such as Adam, AdamW, AdaBelief, and Adan in accuracy, robustness, and convergence speed. Our detailed ablation and sensitivity analyses underscored the crucial role of adaptive learning rate scaling and bias-corrected moment estimation, along with hyperparameter tuning ($\beta_1, \beta_2, \eta_0$). These findings solidify LyAm's potential as a broadly applicable optimization tool and pave the way for future exploration into other complex optimization scenarios.
%$$$$$$$$$$$$$$$$$$$$$$$$$$$$$$$$$$$$$$$$$$$$$$$$$$$$$$$$$$$$$$$$$$$$$$$$$$$$$$$$$$$$$$$$$$$
%$$$$$$$$$$$$$$$$$$$$$$$$$$$$$$$$$$$$$$$$$$$$$$$$$$$$$$$$$$$$$$$$$$$$$$$$$$$$$$$$$$$$$$$$$$$
{
\bibliographystyle{plainnat}
\bibliography{main}

\begin{thebibliography}{52}
\providecommand{\natexlab}[1]{#1}
\providecommand{\url}[1]{\texttt{#1}}
\expandafter\ifx\csname urlstyle\endcsname\relax
  \providecommand{\doi}[1]{doi: #1}\else
  \providecommand{\doi}{doi: \begingroup \urlstyle{rm}\Url}\fi

\bibitem[Bottou et~al.(2018)Bottou, Curtis, and Nocedal]{bottou2018optnet}
L{\'e}on Bottou, Frank~E Curtis, and Jorge Nocedal.
\newblock Optimization methods for large-scale machine learning.
\newblock In \emph{SIAM Review}, volume~60, pages 223--311, 2018.

\bibitem[Chaudhari and et~al.(2019)]{chaudhari2019robust}
Pratik Chaudhari and et~al.
\newblock Robust optimization for deep learning.
\newblock \emph{Journal of Machine Learning Research}, 20:\penalty0 1--42,
  2019.

\bibitem[Dosovitskiy et~al.(2020)Dosovitskiy, Beyer, Kolesnikov, Weissenborn,
  Zhai, Unterthiner, Dehghani, Minderer, Heigold, Gelly,
  et~al.]{dosovitskiy2020image}
Alexey Dosovitskiy, Lucas Beyer, Alexander Kolesnikov, Dirk Weissenborn,
  Xiaohua Zhai, Thomas Unterthiner, Mostafa Dehghani, Matthias Minderer, Georg
  Heigold, Sylvain Gelly, et~al.
\newblock An image is worth 16x16 words: Transformers for image recognition at
  scale.
\newblock \emph{arXiv preprint arXiv:2010.11929}, 2020.

\bibitem[Duchi et~al.(2011)Duchi, Hazan, and Singer]{duchi2011adaptive}
John Duchi, Elad Hazan, and Yoram Singer.
\newblock Adaptive subgradient methods for online learning and stochastic
  optimization.
\newblock \emph{Journal of machine learning research}, 12\penalty0 (7), 2011.

\bibitem[Girshick(2015)]{girshick2015fast}
Ross Girshick.
\newblock Fast r-cnn.
\newblock \emph{IEEE International Conference on Computer Vision (ICCV)}, pages
  1440--1448, 2015.

\bibitem[Goodfellow et~al.(2016)Goodfellow, Bengio, and
  Courville]{goodfellow2016deeplearning}
Ian Goodfellow, Yoshua Bengio, and Aaron Courville.
\newblock \emph{Deep Learning}.
\newblock MIT press, Cambridge, MA, 2016.

\bibitem[Hassan et~al.(2020)Hassan, Hassan, Huda, and
  de~Albuquerque]{hassan2020robust}
Mohammad~Mehedi Hassan, Md~Rafiul Hassan, Shamsul Huda, and Victor Hugo~C
  de~Albuquerque.
\newblock A robust deep-learning-enabled trust-boundary protection for
  adversarial industrial iot environment.
\newblock \emph{IEEE Internet of Things Journal}, 8\penalty0 (12):\penalty0
  9611--9621, 2020.

\bibitem[He et~al.(2016)He, Zhang, Ren, and Sun]{he2016deep}
Kaiming He, Xiangyu Zhang, Shaoqing Ren, and Jian Sun.
\newblock Deep residual learning for image recognition.
\newblock In \emph{Proceedings of the IEEE Conference on Computer Vision and
  Pattern Recognition (CVPR)}, pages 770--778, 2016.

\bibitem[Huk(2020)]{huk2020stochastic}
Maciej Huk.
\newblock Stochastic optimization of contextual neural networks with rmsprop.
\newblock In \emph{Intelligent Information and Database Systems: 12th Asian
  Conference, ACIIDS 2020, Phuket, Thailand, March 23--26, 2020, Proceedings,
  Part II 12}, pages 343--352. Springer, 2020.

\bibitem[Jiang(2022)]{jiang2022security}
Guiwen Jiang.
\newblock Security detection design for laboratory networks based on enhanced
  lstm and adamw algorithms.
\newblock \emph{International Journal of Information Technologies and Systems
  Approach (IJITSA)}, 16\penalty0 (2):\penalty0 1--13, 2022.

\bibitem[Kang et~al.(2019)Kang, Pan, Hoi, and Xu]{kang2019robust}
Zhao Kang, Haiqi Pan, Steven~CH Hoi, and Zenglin Xu.
\newblock Robust graph learning from noisy data.
\newblock \emph{IEEE transactions on cybernetics}, 50\penalty0 (5):\penalty0
  1833--1843, 2019.

\bibitem[Karami et~al.(2023)Karami, Rezaee, Mirzabeigi, and
  Parand]{karami2023comparison}
Amir~Hossein Karami, Sepehr Rezaee, Elmira Mirzabeigi, and Kourosh Parand.
\newblock Comparison of pre-training and classification models for early
  detection of alzheimer’s disease using magnetic resonance imaging.
\newblock In \emph{8th International Conference on Combinatorics Cryptography,
  Computer Science and Computation}, 2023.

\bibitem[Kingma and Ba(2014)]{kingma2014adam}
Diederik~P Kingma and Jimmy Ba.
\newblock Adam: A method for stochastic optimization.
\newblock \emph{arXiv preprint arXiv:1412.6980}, 2014.

\bibitem[Krizhevsky and Nair(2012)]{krizhevsky2012cifar}
Alex Krizhevsky and Vinod Nair.
\newblock Cifar-100 (canadian institute for advanced research). 30 [65] alex
  krizhevsky, ilya sutskever, and geoffrey e hinton. imagenet classification
  with deep convolutional neural networks.
\newblock \emph{Advances in neural information processing systems}, 25\penalty0
  (1097-1105):\penalty0 26, 2012.

\bibitem[Krizhevsky et~al.(2010)Krizhevsky, Nair, and
  Hinton]{krizhevsky2010cifar}
Alex Krizhevsky, Vinod Nair, and Geoffrey Hinton.
\newblock Cifar-10 (canadian institute for advanced research).
\newblock \emph{URL http://www. cs. toronto. edu/kriz/cifar. html}, 5\penalty0
  (4):\penalty0 1, 2010.

\bibitem[Krizhevsky et~al.(2012)Krizhevsky, Sutskever, and
  Hinton]{krizhevsky2012imagenet}
Alex Krizhevsky, Ilya Sutskever, and Geoffrey~E Hinton.
\newblock Imagenet classification with deep convolutional neural networks.
\newblock \emph{Advances in neural information processing systems}, 25, 2012.

\bibitem[Le and Yang(2015)]{le2015tiny}
Yann Le and Xuan Yang.
\newblock Tiny imagenet visual recognition challenge.
\newblock \emph{CS 231N}, 7\penalty0 (7):\penalty0 3, 2015.

\bibitem[Lechner et~al.(2023)Lechner, Amini, Rus, and
  Henzinger]{lechner2023revisiting}
Mathias Lechner, Alexander Amini, Daniela Rus, and Thomas~A Henzinger.
\newblock Revisiting the adversarial robustness-accuracy tradeoff in robot
  learning.
\newblock \emph{IEEE Robotics and Automation Letters}, 8\penalty0 (3):\penalty0
  1595--1602, 2023.

\bibitem[Li et~al.(2020)Li, Socher, and Hoi]{li2020robust}
Xiaobo Li, Richard Socher, and Steven~CH Hoi.
\newblock Robust training under label noise by over-parameterization.
\newblock In \emph{International Conference on Machine Learning (ICML)}, 2020.

\bibitem[Liu et~al.(2020)Liu, Chen, Denil, and et~al.]{liu2020understanding}
Jiajun Liu, Ting-Hsuan Chen, Misha Denil, and et~al.
\newblock Understanding the generalization benefit of weight decay in adam.
\newblock \emph{arXiv preprint arXiv:2006.07358}, 2020.

\bibitem[Liu et~al.(2022)Liu, Hu, Shao, Wang, Zhu, Cui, Wang, Luo, and
  Chen]{liu2022adan}
Zhuangzhuang Liu, Shiji Hu, Wei Shao, Jianyuan Wang, Yifan Zhu, Bin Cui,
  Xiaolin Wang, Ping Luo, and Xiaogang Chen.
\newblock Adan: Adaptive nesterov momentum algorithm for faster optimizing deep
  models.
\newblock In \emph{NeurIPS}, 2022.

\bibitem[Long et~al.(2015)Long, Shelhamer, and Darrell]{long2015fully}
Jonathan Long, Evan Shelhamer, and Trevor Darrell.
\newblock Fully convolutional networks for semantic segmentation.
\newblock In \emph{IEEE Conference on Computer Vision and Pattern Recognition
  (CVPR)}, pages 3431--3440, 2015.

\bibitem[Loshchilov and Hutter(2019)]{loshchilov2019decoupled}
Ilya Loshchilov and Frank Hutter.
\newblock Decoupled weight decay regularization.
\newblock \emph{International Conference on Learning Representations (ICLR)},
  2019.

\bibitem[Lu et~al.(2023)Lu, Li, and Qiu]{lu2023adamw}
Zhuoer Lu, Xiaoyong Li, and Pengfei Qiu.
\newblock An adamw-based deep neural network using feature selection and data
  oversampling for intrusion detection.
\newblock In \emph{2023 8th International Conference on Data Science in
  Cyberspace (DSC)}, pages 213--220. IEEE, 2023.

\bibitem[Madry et~al.(2018)Madry, Makelov, Schmidt, Tsipras, and
  Vladu]{madry2018towards}
Aleksander Madry, Aleksandar Makelov, Ludwig Schmidt, Dimitris Tsipras, and
  Adrian Vladu.
\newblock Towards deep learning models resistant to adversarial attacks.
\newblock In \emph{International Conference on Learning Representations
  (ICLR)}, 2018.

\bibitem[Naifar et~al.(2016)Naifar, Makhlouf, and Hammami]{naifar2016comments}
Omar Naifar, Abdellatif~Ben Makhlouf, and Mohamed~Ali Hammami.
\newblock Comments on “lyapunov stability theorem about fractional system
  without and with delay”.
\newblock \emph{Communications in Nonlinear Science and Numerical Simulation},
  30\penalty0 (1-3):\penalty0 360--361, 2016.

\bibitem[Oikarinen et~al.(2021)Oikarinen, Zhang, Megretski, Daniel, and
  Weng]{oikarinen2021robust}
Tuomas Oikarinen, Wang Zhang, Alexandre Megretski, Luca Daniel, and Tsui-Wei
  Weng.
\newblock Robust deep reinforcement learning through adversarial loss.
\newblock \emph{Advances in Neural Information Processing Systems},
  34:\penalty0 26156--26167, 2021.

\bibitem[Patrini et~al.(2017)Patrini, Rozza, Krishna~Menon, Nock, and
  Qu]{patrini2017making}
Giorgio Patrini, Alessandro Rozza, Aditya Krishna~Menon, Richard Nock, and
  Lizhen Qu.
\newblock Making deep neural networks robust to label noise: A loss correction
  approach.
\newblock \emph{CVPR}, 2017.

\bibitem[Pattanaik et~al.(2017)Pattanaik, Tang, Liu, Bommannan, and
  Chowdhary]{pattanaik2017robust}
Anay Pattanaik, Zhenyi Tang, Shuijing Liu, Gautham Bommannan, and Girish
  Chowdhary.
\newblock Robust deep reinforcement learning with adversarial attacks.
\newblock \emph{arXiv preprint arXiv:1712.03632}, 2017.

\bibitem[Polyak(1964)]{polyak1964some}
Boris~T Polyak.
\newblock Some methods of speeding up the convergence of iteration methods.
\newblock \emph{Ussr computational mathematics and mathematical physics},
  4\penalty0 (5):\penalty0 1--17, 1964.

\bibitem[Reddi et~al.(2018)Reddi, Kale, and Kumar]{reddi2018convergence}
Sashank~J Reddi, Satyen Kale, and Sanjiv Kumar.
\newblock On the convergence of adam and beyond.
\newblock In \emph{International Conference on Learning Representations
  (ICLR)}, 2018.

\bibitem[Ren et~al.(2015)Ren, He, Girshick, and Sun]{ren2015faster}
Shaoqing Ren, Kaiming He, Ross Girshick, and Jian Sun.
\newblock Faster r-cnn: Towards real-time object detection with region proposal
  networks.
\newblock In \emph{Advances in Neural Information Processing Systems}, pages
  91--99, 2015.

\bibitem[Robbins and Monro(1951)]{robbins1951stochastic}
Herbert Robbins and Sutton Monro.
\newblock A stochastic approximation method.
\newblock \emph{The annals of mathematical statistics}, pages 400--407, 1951.

\bibitem[Rolnick et~al.(2017)Rolnick, Veit, Belongie, and
  Shavit]{rolnick2017deep}
David Rolnick, Andreas Veit, Serge Belongie, and Nir Shavit.
\newblock Deep learning is robust to massive label noise.
\newblock \emph{arXiv preprint arXiv:1705.10694}, 2017.

\bibitem[Roy et~al.(2018)Roy, Ahmed, and Akhand]{roy2018noisy}
Sudipta~Singha Roy, Mahtab Ahmed, and Muhammad Aminul~Haque Akhand.
\newblock Noisy image classification using hybrid deep learning methods.
\newblock \emph{Journal of Information and Communication Technology},
  17\penalty0 (2):\penalty0 233--269, 2018.

\bibitem[Sastry and Sastry(1999)]{sastry1999lyapunov}
Shankar Sastry and Shankar Sastry.
\newblock Lyapunov stability theory.
\newblock \emph{Nonlinear systems: analysis, stability, and control}, pages
  182--234, 1999.

\bibitem[Shevitz and Paden(1994)]{shevitz1994lyapunov}
Daniel Shevitz and Brad Paden.
\newblock Lyapunov stability theory of nonsmooth systems.
\newblock \emph{IEEE Transactions on automatic control}, 39\penalty0
  (9):\penalty0 1910--1914, 1994.

\bibitem[Simonyan and Zisserman(2014)]{simonyan2014very}
Karen Simonyan and Andrew Zisserman.
\newblock Very deep convolutional networks for large-scale image recognition.
\newblock In \emph{International Conference on Learning Representations
  (ICLR)}, 2014.

\bibitem[Stallkamp et~al.(2011)Stallkamp, Schlipsing, Salmen, and
  Igel]{stallkamp2011german}
Johannes Stallkamp, Marc Schlipsing, Jan Salmen, and Christian Igel.
\newblock The german traffic sign recognition benchmark: a multi-class
  classification competition.
\newblock In \emph{The 2011 international joint conference on neural networks},
  pages 1453--1460. IEEE, 2011.

\bibitem[Sukhbaatar et~al.(2014)Sukhbaatar, Bruna, Paluri, Bourdev, and
  Fergus]{sukhbaatar2014training}
Sainbayar Sukhbaatar, Joan Bruna, Manohar Paluri, Lubomir Bourdev, and Rob
  Fergus.
\newblock Training convolutional networks with noisy labels.
\newblock \emph{arXiv preprint arXiv:1406.2080}, 2014.

\bibitem[Szegedy et~al.(2014)Szegedy, Zaremba, Sutskever, Bruna, Erhan,
  Goodfellow, and Fergus]{szegedy2013intriguing}
Christian Szegedy, Wojciech Zaremba, Ilya Sutskever, Joan Bruna, Dumitru Erhan,
  Ian Goodfellow, and Rob Fergus.
\newblock Intriguing properties of neural networks.
\newblock \emph{International Conference on Learning Representations (ICLR)},
  2014.

\bibitem[Tamkin et~al.(2020)Tamkin, Yong, Manning, and Goodman]{tamkin2020view}
Aaron Tamkin, Hanlin Yong, Christopher~D Manning, and Noah~D Goodman.
\newblock Viewmaker networks: Learning views for unsupervised representation
  learning.
\newblock \emph{arXiv preprint arXiv:2010.07432}, 2020.

\bibitem[Tsipras et~al.(2018)Tsipras, Santurkar, Engstrom, Turner, and
  Madry]{tsipras2018robustness}
Dimitris Tsipras, Shibani Santurkar, Logan Engstrom, Alexander Turner, and
  Aleksander Madry.
\newblock Robustness may be at odds with accuracy.
\newblock \emph{arXiv preprint arXiv:1805.12152}, 2018.

\bibitem[Xu et~al.(2015)Xu, Ba, Kiros, Cho, Courville, Salakhudinov, Zemel, and
  Bengio]{xu2015show}
Kelvin Xu, Jimmy Ba, Ryan Kiros, Kyunghyun Cho, Aaron Courville, Ruslan
  Salakhudinov, Rich Zemel, and Yoshua Bengio.
\newblock Show, attend and tell: Neural image caption generation with visual
  attention.
\newblock In \emph{International conference on machine learning}, pages
  2048--2057. PMLR, 2015.

\bibitem[Yang et~al.(2023)Yang, Zhang, Li, Zhang, Xu, and
  Wang]{yang2023adversarial}
Bo~Yang, Hengwei Zhang, Zheming Li, Yuchen Zhang, Kaiyong Xu, and Jindong Wang.
\newblock Adversarial example generation with adabelief optimizer and crop
  invariance.
\newblock \emph{Applied Intelligence}, 53\penalty0 (2):\penalty0 2332--2347,
  2023.

\bibitem[Yoshida(2024)]{yoshida2024learning}
Soh Yoshida.
\newblock Learning with noisy labels for image classification.
\newblock \emph{IEICE ESS Fundamentals Review}, 18\penalty0 (2):\penalty0
  147--157, 2024.

\bibitem[Zaheer et~al.(2018)Zaheer, Reddi, Sachan, Kale, and
  Kumar]{zaheer2018adaptive}
Manzil Zaheer, Sashank Reddi, Devendra Sachan, Satyen Kale, and Sanjiv Kumar.
\newblock Adaptive methods for nonconvex optimization.
\newblock \emph{Advances in neural information processing systems}, 31, 2018.

\bibitem[Zhang et~al.(2016)Zhang, Bengio, Hardt, Recht, and
  Vinyals]{zhang2016understanding}
Chiyuan Zhang, Samy Bengio, Moritz Hardt, Benjamin Recht, and Oriol Vinyals.
\newblock Understanding deep learning requires rethinking generalization.
\newblock \emph{arXiv preprint arXiv:1611.03530}, 2016.

\bibitem[Zhang et~al.(2017)Zhang, Bengio, Hardt, Recht, and
  Vinyals]{zhang2017understanding}
Chiyuan Zhang, Samy Bengio, Moritz Hardt, Benjamin Recht, and Oriol Vinyals.
\newblock Understanding deep learning requires rethinking generalization.
\newblock \emph{International Conference on Learning Representations (ICLR)},
  2017.

\bibitem[Zhang et~al.(2023)Zhang, Hu, Xu, and Zhao]{zhang2023yolov7}
Chong Zhang, Zhuhua Hu, Lewei Xu, and Yaochi Zhao.
\newblock A yolov7 incorporating the adan optimizer based corn pests
  identification method.
\newblock \emph{Frontiers in Plant Science}, 14:\penalty0 1174556, 2023.

\bibitem[Zhu et~al.(2017)Zhu, Park, Isola, and Efros]{zhu2017unpaired}
Jun-Yan Zhu, Taesung Park, Phillip Isola, and Alexei~A Efros.
\newblock Unpaired image-to-image translation using cycle-consistent
  adversarial networks.
\newblock In \emph{Proceedings of the IEEE international conference on computer
  vision}, pages 2223--2232, 2017.

\bibitem[Zhuang et~al.(2020)Zhuang, Tang, Ding, Tatikonda, Dvornek,
  Papademetris, and Duncan]{zhuang2020adabelief}
Juntang Zhuang, Tommy Tang, Yifan Ding, Sekhar~C Tatikonda, Nicha Dvornek,
  Xenophon Papademetris, and James Duncan.
\newblock Adabelief optimizer: Adapting stepsizes by the belief in observed
  gradients.
\newblock \emph{Advances in neural information processing systems},
  33:\penalty0 18795--18806, 2020.

\end{thebibliography}
}
%$$$$$$$$$$$$$$$$$$$$$$$$$$$$$$$$$$$$$$$$$$$$$$$$$$$$$$$$$$$$$$$$$$$$$$$$$$$$$$$$$$$$$$$$$$$
%$$$$$$$$$$$$$$$$$$$$$$$$$$$$$$$$$$$$$$$$$$$$$$$$$$$$$$$$$$$$$$$$$$$$$$$$$$$$$$$$$$$$$$$$$$$
\end{document}